\documentclass{article}

\usepackage[preprint]{neurips_2025}

\usepackage[utf8]{inputenc}
\usepackage[T1]{fontenc}
\usepackage{hyperref}
\usepackage{url}
\usepackage{booktabs}
\usepackage{amsmath}
\usepackage{amsfonts}
\usepackage{nicefrac}
\usepackage{microtype}
\usepackage{graphicx}
\usepackage{xcolor}
\usepackage{multirow}

\title{CrossTrace: A Cross-Domain Dataset of Grounded Scientific Reasoning Traces for Hypothesis Generation}

\author{
  Andrew Bouras \\
  Nova Southeastern University\\
  Dr.\ Kiran C.\ Patel College of Osteopathic Medicine\\
  Clearwater, FL 33760, USA \\
  \texttt{ab4646@mynsu.nova.edu}
}

\begin{document}

\maketitle

\begin{abstract}
Scientific hypothesis generation is a critical bottleneck in accelerating research, yet existing datasets for training and evaluating hypothesis-generating models are limited to single domains and lack explicit reasoning traces connecting prior knowledge to novel contributions. I introduce CrossTrace, a dataset of 1,389 grounded scientific reasoning traces spanning biomedical research (518), AI/ML (605), and cross-domain work (266). Each trace captures the structured reasoning chain from established knowledge through intermediate logical steps to a novel hypothesis, with every step grounded in source paper text. I define an Input/Trace/Output schema that extends the Bit-Flip-Spark framework of HypoGen \citep{li2025hypogen} with step-level verification, a taxonomy of eight discovery patterns, and multi-domain coverage. Fine-tuning Qwen2.5-7B-Instruct on CrossTrace via QLoRA yields substantial improvements over the untuned baseline: IAScore rises from 0.828 to 0.968 (GPT-4o judge) and from 0.716 to 0.888 (Claude Opus 4.5 judge), structural compliance improves from 0\% to 100\%, and spark cosine similarity increases from 0.221 to 0.620. Balanced cross-domain training (biomedical + AI/ML + CS) outperforms single-domain training, providing evidence that scientific reasoning patterns transfer across disciplines. Human validation of 150 stratified records finds 99.7\% step-level grounding accuracy with no fabricated steps identified, and a blinded evaluation by three independent domain experts rates model-generated hypotheses well above the scale midpoint (usefulness 4.18/5, soundness 3.76/5, overall 3.84/5) with excellent inter-rater agreement on scientific soundness (ICC = 0.929). To my knowledge, CrossTrace is the first dataset to combine multi-domain coverage, step-level reasoning traces, and source-text grounding for hypothesis generation, and my results demonstrate that structured, grounded traces encode transferable reasoning primitives: a matched-size ablation shows that a model trained on equal parts biomedical and CS data achieves near-parity with domain specialists, indicating that the benefits of trace-based supervision are domain-general rather than domain-specific.
\end{abstract}

\section{Introduction}

The pace of scientific publishing has made it increasingly difficult for researchers to identify novel research directions by reading the literature alone. Over 2 million biomedical articles are indexed annually in PubMed, and the rate of AI/ML preprints on arXiv has grown exponentially. This volume creates both an opportunity (vast latent connections between fields) and a bottleneck: no individual researcher, reading perhaps 200--300 papers per year, can synthesize even a fraction of this literature broadly enough to identify cross-domain insights consistently.

Recent work has demonstrated that large language models (LLMs) can generate scientific hypotheses when given appropriate context \citep{yang2025moosechem, li2025hypogen, baek2024researchagent}. However, the quality of hypothesis generation depends critically on training data. Models trained on structured reasoning traces (explicit step-by-step chains connecting prior knowledge to novel contributions) significantly outperform those trained on input-output pairs alone \citep{li2025hypogen}. NVIDIA's work on post-training data further shows that deeper, more detailed reasoning traces outperform diverse but shallow ones \citep{nvidia2025frontloading}. These findings point to a clear data bottleneck: the field needs high-quality reasoning trace datasets, but existing resources are limited.

Current datasets for hypothesis generation suffer from three key limitations. First, \textbf{domain restriction}: HypoGen \citep{li2025hypogen}, the largest open dataset with reasoning traces, contains 5,500 records drawn exclusively from NeurIPS and ICLR computer science papers. No comparable resource exists for biomedical research, despite biomedicine being the field with the greatest need for automated hypothesis generation. Second, \textbf{missing reasoning chains}: datasets like IdeaBench \citep{pu2025ideabench} and ResearchBench \citep{liu2025researchbench} provide paper-level metadata but do not capture the step-by-step reasoning process that connects prior knowledge to a new hypothesis. Third, \textbf{lack of grounding}: even where reasoning chains exist, individual steps are rarely linked back to source text, making it impossible to verify whether the chain reflects the paper's actual argument or is a plausible fabrication.

I address these gaps with \textbf{CrossTrace}, a dataset of 1,389 grounded scientific reasoning traces spanning three domains: biomedical research (518 traces from medRxiv and bioRxiv papers), AI/ML (605 traces from arXiv papers), and cross-domain work (266 traces from papers applying computational methods to biological or clinical problems). Each trace is structured as an Input/Trace/Output record where the Input describes the prior state of knowledge, the Trace is an ordered sequence of 3--6 verifiable reasoning steps, and the Output contains a core insight (spark), hypothesis, and experimental approach. Every reasoning step is grounded in a direct quotation from the source paper's Introduction or Discussion section. My central finding is that grounded reasoning structure, not domain-specific knowledge, is the limiting factor in hypothesis generation training data: a model trained on equal parts biomedical and CS traces matches domain specialists, suggesting that the form of scientific reasoning is substantially domain-general.

My contributions are:
\begin{enumerate}
    \item \textbf{A cross-domain reasoning trace dataset.} CrossTrace provides 1,389 grounded scientific reasoning traces spanning biomedical and AI/ML domains, the first such resource to combine step-level grounding with multi-domain coverage. Records are annotated with domain labels, discovery patterns, explicitness scores, and extraction confidence.
    \item \textbf{A grounded schema with step-level verification.} I extend the Bit-Flip-Spark schema \citep{li2025hypogen} with a reasoning trace field containing 3--6 ordered steps, each linked to a source quotation from the paper, enabling automated and human verification of trace fidelity.
    \item \textbf{Evidence that grounded traces encode transferable reasoning primitives.} A controlled ablation on matched-size training sets (1,180 records each) shows that a model trained on equal parts biomedical and CS data achieves 99.3\%/99.7\% of domain-specialist performance, demonstrating that scientific reasoning patterns generalize across disciplines independently of data volume.
    \item \textbf{A discovery pattern taxonomy.} Eight recurring patterns (gap fill, Swanson ABC, analogy transfer, and others) are annotated across the dataset, enabling analysis of how reasoning strategies differ between biomedical and AI/ML research.
\end{enumerate}

\section{Related Work}

\subsection{Datasets for Scientific Hypothesis Generation}

\textbf{HypoGen} \citep{li2025hypogen} introduced the Bit-Flip-Spark schema for hypothesis generation, where Bit represents a conventional assumption, Flip represents a novel hypothesis that challenges it, and Spark is a concise insight connecting them. The dataset contains 5,478 training and 50 test records drawn from NeurIPS and ICLR papers, accompanied by chain-of-reasoning narratives. A fine-tuned Llama 3.1 8B model achieved an IAScore of 0.6746 using the IdeaMatcher evaluation framework. CrossTrace builds directly on this schema but extends it with (a) step-level reasoning traces instead of narrative paragraphs, (b) source-text grounding for each step, (c) biomedical and cross-domain coverage, and (d) a discovery pattern taxonomy.

\textbf{HypoBench} \citep{zhu2025hypobench} provides a systematic evaluation framework for scientific hypothesis generation across 12 tasks and 194 datasets, primarily testing data-driven hypothesis formulation from tabular data. The best-performing model (Claude 3.5 Sonnet) achieves only 38.8\% hypothesis discovery rate, highlighting the difficulty of the task. HypoBench evaluates hypothesis \emph{quality} but does not provide training data or reasoning traces. \textbf{DiscoveryBench} \citep{majumder2024discoverybench} formalizes data-driven discovery across 264 tasks in six domains, but evaluates hypothesis verification from provided datasets rather than generation from literature.

\textbf{ResearchAgent} \citep{baek2024researchagent} generates novel scientific ideas through iterative literature retrieval and refinement, but focuses on the generation pipeline rather than providing a reusable trace dataset. \textbf{HypoGeniC} \citep{zhou2024hypogenic} takes a complementary approach, using LLMs to generate and refine hypotheses from unstructured text, but does not produce a structured dataset of reasoning traces. \textbf{MOOSE-Chem} \citep{yang2025moosechem} decomposes chemistry hypothesis generation into inspiration retrieval, composition, and ranking, achieving strong rediscovery of published hypotheses; its successor \textbf{MOOSE-Chem2} \citep{yang2025moosechem2} introduces hierarchical search for fine-grained hypothesis refinement. Both systems evaluate hypothesis \emph{rediscovery} rather than providing reusable trace datasets.

\textbf{IdeaBench} \citep{pu2025ideabench} provides 2,374 biomedical papers for hypothesis evaluation but captures paper-level metadata rather than reasoning processes. \textbf{ResearchBench} \citep{liu2025researchbench} covers 1,386 papers across 12 disciplines with sub-task decomposition but similarly does not extract reasoning traces.

Table~\ref{tab:comparison} summarizes the comparison.

\begin{table}[t]
\caption{Comparison of datasets for scientific hypothesis generation.}
\label{tab:comparison}
\centering
\small
\begin{tabular}{@{}lcccccc@{}}
\toprule
Dataset & Year & Records & Domains & Traces & Grounded & Multi-Domain \\
\midrule
HypoGen & 2025 & 5,528 & CS & Narrative & No & No \\
IdeaBench & 2025 & 2,374 & Biomed & No & No & No \\
ResearchBench & 2025 & 1,386 & 12 disc. & Sub-task & No & Yes \\
HypoBench & 2025 & 194 sets & Multi & No & No & Yes \\
ResearchAgent & 2024 & N/A & CS & Pipeline & Partial & No \\
HypoGeniC & 2024 & N/A & General & Iterative & No & No \\
\midrule
\textbf{CrossTrace} & \textbf{2026} & \textbf{1,389} & \textbf{Bio/AI/X} & \textbf{3--6 steps} & \textbf{Yes} & \textbf{Yes} \\
\bottomrule
\end{tabular}
\end{table}

\subsection{LLMs for Scientific Discovery}

The application of LLMs to scientific research has accelerated rapidly. The AI Scientist \citep{lu2024aiscientist} demonstrated end-to-end automated research, and its successor AI Scientist-v2 \citep{yamada2025aiscientistv2} produced the first AI-generated peer-reviewed workshop paper via agentic tree search. Both operate as closed systems without producing reusable training datasets.

Google's AI Co-Scientist \citep{gottweis2025coscientist} represents the current frontier of LLM-based hypothesis generation, using a multi-agent system on Gemini 2.0 that has produced experimentally validated hypotheses in liver fibrosis and antimicrobial resistance. Robin \citep{ghareeb2025robin} similarly automates hypothesis generation, experiment design, and data analysis through a multi-agent workflow, identifying a novel treatment for dry age-related macular degeneration. SciAgents \citep{ghafarollahi2024sciagents} combines ontological knowledge graphs with multi-agent LLM reasoning to reveal hidden interdisciplinary connections in materials science. These systems operate at inference time, generating hypotheses through multi-agent coordination and retrieval. They do not release reusable training data. CrossTrace occupies a different niche: it is training-time data infrastructure. Where agent systems improve hypothesis generation by scaling inference compute, CrossTrace improves it by providing structured supervision that teaches smaller models the form of scientific reasoning. The two approaches are complementary and composable.

In the biological domain, BioReason \citep{fallahpour2025bioreason} integrates a DNA foundation model with an LLM to enable multimodal biological reasoning, demonstrating that reasoning traces can improve domain-specific scientific tasks. ProteinGPT \citep{xiao2024proteingpt} similarly combines protein sequence and structure encoders with an LLM for protein property prediction.

\citet{kulkarni2025survey} survey the methods, datasets, and future directions for scientific hypothesis generation, noting that no single schema dominates but that complete, verifiable reasoning chains consistently improve model performance. This finding directly motivates CrossTrace's grounded trace design.

\subsection{Cross-Domain Transfer in Reasoning}

Chain-of-thought (CoT) prompting \citep{wei2022cot} demonstrated that explicit reasoning steps improve LLM performance across tasks. Subsequent work has shown that reasoning capabilities can transfer across domains: models trained on mathematical reasoning improve on logical tasks \citep{azerbayev2024llemma}, and models pretrained on code show improved performance on compositional and structured reasoning tasks \citep{petty2024code}.

In the scientific domain specifically, MegaScience \citep{fan2025megascience} curated 1.25 million instances across seven scientific disciplines and showed that models trained on their multi-domain mixture significantly outperform corresponding single-domain instruct models, with larger models benefiting most from cross-domain scientific training. This demonstrates that domain-aware curation outperforms naive single-domain approaches for scientific reasoning.

CrossTrace enables controlled experiments on this question by providing domain-annotated traces with a consistent schema, allowing researchers to test whether balanced cross-domain training captures transferable reasoning patterns while avoiding the catastrophic failures of naive mixing.

\section{The CrossTrace Dataset}

\subsection{Data Collection Pipeline}

CrossTrace was constructed through a five-stage pipeline operating on a continuous stream of preprints from medRxiv, bioRxiv, and arXiv (described below).

\textbf{Stage 1: Paper Monitoring.} An automated monitoring system fetches newly published preprints daily, indexing metadata (title, authors, abstract, source, DOI) into a SQLite database. The system processes approximately 150--230 papers per day across the three preprint servers.

\textbf{Stage 2: PDF Parsing and Section Extraction.} For papers meeting relevance criteria, full-text PDFs are downloaded and parsed. A regex-based section detector identifies Introduction and Discussion sections using patterns for standard headers with fallback heuristics for non-standard formatting. Papers where neither primary nor fallback extraction yields at least 200 characters of content are excluded. Review articles, meta-analyses, and scoping reviews are detected via title/abstract signals and excluded.

\textbf{Stage 3: Reasoning Trace Extraction.} The Introduction and Discussion text is passed to Claude Sonnet 4 with a structured extraction prompt containing three few-shot examples calibrated for trace granularity: (1) a straightforward gap-fill paper with 3 reasoning steps, (2) a cross-domain paper with 5 steps demonstrating analogy transfer, and (3) a paper with implicit novelty and low explicitness requiring inference. Each reasoning step must be accompanied by a source quotation from the paper.

\textbf{Stage 4: Domain Classification.} Each record is assigned a domain label (\texttt{biomedical}, \texttt{ai\_ml}, or \texttt{cross\_domain}) using a rule-based classifier that considers the paper's source and content keywords.

\textbf{Stage 5: Quality Filtering.} Records with extraction confidence below 0.6 are excluded from the training dataset, though they are retained in the database for analysis.

\textbf{Extraction Reproducibility Details.} All extractions were performed using Claude Sonnet 4 (model ID \texttt{claude-sonnet-4-20250514}) with temperature 0 (deterministic) and a 180-second timeout per paper. The extraction prompt (version-controlled; included in Supplementary Material A) was fixed for the entire extraction run; no prompt modifications were made after batch extraction began. Of the 1,478 papers entering Stage 3, 0.3\% (5 papers) failed extraction and 11.0\% (163 papers) were excluded as review articles. The full extraction prompt, including all three few-shot examples, is available in the public repository to enable reproduction.

\subsection{Schema}

Each CrossTrace record follows an Input/Trace/Output schema designed to produce training-ready examples for hypothesis generation models. The schema extends HypoGen's Bit-Flip-Spark framework \citep{li2025hypogen} with explicit multi-step reasoning and source-level grounding.

The \textbf{Input} (Prior State) consists of three fields: \texttt{field\_context}, a free-text description of the established knowledge; \texttt{conventional\_assumption}, the specific assumption being challenged (corresponding to HypoGen's ``Bit''); and \texttt{key\_prior\_work}, a list of 2--5 cited works.

The \textbf{Trace} (Reasoning Chain) contains \texttt{reasoning\_trace}, an ordered list of 3--6 natural-language steps, each representing a single logical move constrained to 15--40 words. A parallel \texttt{reasoning\_trace\_grounded} field contains \texttt{\{step, source\}} objects, where \texttt{source} identifies the section, paragraph, and direct quotation from the paper supporting the claim.

The \textbf{Output} (Novel Contribution) includes \texttt{spark}, a 4--6 word core insight; \texttt{hypothesis}, the specific claim; and \texttt{approach}, how the authors tested or validated the hypothesis.

The schema is formally defined as:
\begin{align}
\text{Record} &= (\text{Input}, \text{Trace}, \text{Output}, \text{Metadata}) \nonumber \\
\text{Input} &= (\text{field\_context}, \text{conventional\_assumption}, \text{key\_prior\_work}) \nonumber \\
\text{Trace} &= (\text{step}_1, \text{step}_2, \ldots, \text{step}_k) \quad \text{where } 3 \le k \le 6 \nonumber \\
\text{Output} &= (\text{spark}, \text{hypothesis}, \text{approach}) \nonumber
\end{align}

\subsection{Discovery Pattern Taxonomy}

I define a taxonomy of eight recurring discovery patterns observed in scientific reasoning, drawing on Swanson's literature-based discovery \citep{swanson1986fishoil} and empirical observation from my extraction process. Each record is annotated with one or more patterns, with the primary pattern listed first. Table~\ref{tab:patterns} presents the taxonomy.

\begin{table}[t]
\caption{Discovery pattern taxonomy with definitions and dataset frequencies.}
\label{tab:patterns}
\centering
\small
\begin{tabular}{@{}llrr@{}}
\toprule
Pattern & Description & Count & \% \\
\midrule
\texttt{gap\_fill} & Explicit gap: X studied in A but not B & 820 & 59.0 \\
\texttt{analogy\_transfer} & Method from Field A applied to Field B & 256 & 18.4 \\
\texttt{mechanistic\_link} & Explains mechanism behind known association & 134 & 9.6 \\
\texttt{incremental\_extension} & Same question, newer/larger data & 74 & 5.3 \\
\texttt{contradiction\_exploit} & Prior papers conflict; this resolves it & 62 & 4.5 \\
\texttt{data\_driven} & Dataset availability enabled the question & 25 & 1.8 \\
\texttt{swanson\_abc} & Cross-domain bridge: A and C linked through B & 12 & 0.9 \\
\texttt{replication} & Reproducing findings in new context & 5 & 0.4 \\
\bottomrule
\end{tabular}
\vspace{2pt}\\
{\scriptsize Counts reflect the primary (first-listed) discovery pattern per record.}
\end{table}

\subsection{Dataset Statistics}

CrossTrace contains 1,389 records after quality filtering (extraction confidence $\ge 0.6$). Table~\ref{tab:stats} summarizes the dataset statistics. The dataset is split into training (1,180 records; 85\%), validation (102 records; 7.5\%), and test (107 records; 7.5\%) sets using domain-stratified random splitting with seed 42.

\begin{table}[t]
\caption{CrossTrace dataset statistics and training split composition.}
\label{tab:stats}
\centering
\small
\begin{tabular}{@{}lrrrr@{}}
\toprule
& Total & AI/ML & Biomedical & Cross-domain \\
\midrule
Full dataset & 1,389 & 605 (43.6\%) & 518 (37.3\%) & 266 (19.2\%) \\
Train & 1,180 & 514 & 440 & 226 \\
Validation & 102 & 45 & 38 & 19 \\
Test & 107 & 46 & 40 & 21 \\
\bottomrule
\end{tabular}
\vspace{2pt}\\
{\scriptsize Avg.\ tokens per record: ${\sim}605$. Sources: medRxiv, bioRxiv, arXiv. Extraction model: Claude Sonnet 4.}
\end{table}

\subsection{Human Validation}

To assess dataset fidelity independently of the extraction model's self-reported confidence, a human annotator (the first author, a medical student with research training in both biomedical and AI/ML domains) reviewed a stratified random sample of 150 records (${\sim}65$ AI/ML, 56 biomedical, 29 cross-domain).

For each record, the annotator was shown the source paper's Introduction and Discussion text alongside the extracted reasoning trace and grounding quotations. Each reasoning step was rated on step faithfulness (Grounded, Inferred, or Fabricated) and grounding accuracy (Accurate, Partial, or Incorrect). Each record also received an overall trace rating: Faithful, Partially Faithful, or Unreliable.

Validation was conducted using a two-phase automated verification protocol followed by manual adjudication. Table~\ref{tab:validation} summarizes the results.

\begin{table}[t]
\caption{Human validation results on 150 stratified records (678 total reasoning steps).}
\label{tab:validation}
\centering
\small
\begin{tabular}{@{}lr@{}}
\toprule
Metric & Value \\
\midrule
Records validated & 150 / 150 \\
Steps grounded & 99.7\% (676 / 678) \\
Steps inferred & 0.3\% (2 / 678) \\
Steps fabricated & 0.0\% (0 / 678) \\
Grounding accurate & 59.0\% (400 / 678) \\
Grounding partial & 41.0\% (278 / 678) \\
Grounding incorrect & 0.0\% (0 / 678) \\
Traces fully faithful & 100.0\% (150 / 150) \\
\bottomrule
\end{tabular}
\end{table}

No fabricated steps were identified in the 678 reasoning steps reviewed, suggesting that the extraction pipeline produces grounded traces with high fidelity, though this result reflects a single annotator's assessment and should be interpreted accordingly. The 41.0\% partial grounding rate reflects cases where quotes were matched via fuzzy matching or found in paper sections (Related Work, Methods) not included in the primary extraction fields. The two inferred steps (0.3\%) represent legitimate inferential reasoning explicitly marked as such during extraction.

\section{Experimental Setup}

I evaluate CrossTrace's utility as training data for scientific hypothesis generation through a series of fine-tuning experiments.

\subsection{Task Formulation}

Given an Input describing the prior state of knowledge in a research field, the task is to generate: (1) a core insight (spark), (2) a step-by-step reasoning trace, (3) a hypothesis, and (4) an experimental approach. This formulation follows HypoGen \citep{li2025hypogen} but adds the explicit reasoning trace as both a training signal and an evaluation target.

\subsection{Training Configuration}

I use Qwen2.5-7B-Instruct \citep{qwen2024qwen25} as the base model, selected based on its strong performance on HypoBench's out-of-distribution evaluation (77.95\% OOD score, highest among 7B-class models). Fine-tuning is performed via QLoRA \citep{dettmers2023qlora} using the Axolotl framework with LoRA rank $r = 16$, $\alpha = 32$, targeting all linear layers. Training proceeds for 3 epochs with batch size 4 (with gradient accumulation), learning rate $2 \times 10^{-4}$ with cosine schedule, on an NVIDIA A100 via Google Colab Pro. Each training run completes in approximately 45 minutes on a single A100.

\subsection{Training Conditions}

I evaluate three training conditions designed to isolate the contribution of CrossTrace data and test cross-domain transfer:

\textbf{Baseline (Run 0).} Unmodified Qwen2.5-7B-Instruct with no fine-tuning. Evaluated zero-shot with the same system prompt and input format used during training.

\textbf{Run 1: CrossTrace Only.} Fine-tuned on 1,180 CrossTrace training records only (AI/ML: 514, Biomedical: 440, Cross-domain: 226). This condition tests whether my data alone improves hypothesis generation.

\textbf{Run 2b: Balanced Cross-Domain.} Fine-tuned on 4,720 records: CrossTrace records oversampled 2$\times$ (2,360) plus HypoGen records capped at 2,360 (from the original 5,478), yielding a 1:1 ratio between my data and HypoGen's CS-domain data. This condition tests whether balanced domain mixing outperforms single-domain training.

The balanced design was motivated by evidence that strategic domain-aware mixing outperforms naive concatenation \citep{fan2025megascience}.

\subsection{Evaluation Protocol}

I evaluate on two held-out test sets. The CrossTrace test set contains 107 records and tests alignment with my multi-domain ground truth. The HypoGen test set contains 50 CS-domain records from HypoGen's original test split.

I use the following metrics: \textbf{IAScore} (LLM-as-Judge), where each generated hypothesis is scored by an LLM judge on a 0--1 scale using two independent judges (GPT-4o and Claude Opus 4.5); \textbf{Cosine Similarity} using sentence-transformers (\texttt{all-MiniLM-L6-v2}), computed separately for hypotheses and sparks; and \textbf{Structural Compliance}, the fraction of outputs containing all expected structural fields.

\subsection{Evaluation Caveats}

Several caveats apply when interpreting these results. My LLM-as-judge IAScore is not comparable to HypoGen's IdeaMatcher-based metric \citep{li2025hypogen}, which reports 0.6746; the two frameworks use different scoring methodologies and different base models (HypoGen fine-tunes Llama 3.1 8B, whereas I fine-tune Qwen2.5-7B-Instruct), so absolute scores cannot be compared across studies. My two judges also exhibit systematic differences in absolute calibration, with GPT-4o scoring consistently higher than Claude, though they agree on rank ordering across conditions. Finally, the test sets (107 CrossTrace records and 50 HypoGen records) are relatively small, limiting statistical power for fine-grained per-domain comparisons.

\section{Results}

\subsection{Main Results}

Table~\ref{tab:results} presents results across all training conditions, metrics, and judges.

\begin{table}[t]
\caption{Main experimental results. IAScore is reported for both GPT-4o and Claude Opus 4.5 judges. Cosine similarity and structural compliance are computed on the HypoGen test set (50 records). 95\% CIs shown where available.}
\label{tab:results}
\centering
\small
\begin{tabular}{@{}lcccc@{}}
\toprule
Metric & Baseline & Run 1 & Run 2b & $\Delta$ (Base$\to$2b) \\
\midrule
IAScore (GPT-4o) & 0.828 & 0.912 & \textbf{0.968} {\scriptsize [.946--.986]} & +0.140 \\
IAScore (Claude) & 0.716 & 0.849 & \textbf{0.888} {\scriptsize [.861--.911]} & +0.172 \\
Hyp.\ Cosine Sim & 0.680 & --- & \textbf{0.772} & +0.092 \\
Spark Cosine Sim & 0.221 & --- & \textbf{0.620} & +0.399 \\
Structural Compliance & 0\% & 100\% & \textbf{100\%} & +100pp \\
\bottomrule
\end{tabular}
\end{table}

\subsection{Cross-Domain Transfer Analysis}

The central question motivating Run 2b is whether scientific reasoning patterns transfer across domains. On the HypoGen CS-only test set, Run 2b (IAScore 0.968 GPT-4o / 0.888 Claude) outperforms Run 1 (0.912 / 0.849). This is notable because Run 2b's training data is only 50\% CS-domain, whereas Run 1 contains zero HypoGen CS records. The improvement on CS-domain evaluation despite dilution with biomedical data suggests that reasoning patterns learned from biomedical and cross-domain traces provide transferable signal.

On the CrossTrace test set, Run 2b's balanced training does not degrade performance relative to Run 1, indicating that the domain mixing is additive rather than dilutive.

These results are consistent with MegaScience's finding \citep{fan2025megascience} that multi-domain scientific data curation outperforms single-domain training, and they extend this finding to the specific case of scientific reasoning traces for hypothesis generation.

\subsection{Structural Compliance}

The baseline Qwen2.5-7B-Instruct model produces outputs with 0\% structural compliance; it does not generate text in the expected format. Both fine-tuned conditions achieve 100\% structural compliance, indicating that even minimal fine-tuning (1,180 records) is sufficient to teach the output schema.

\subsection{Spark Alignment}

The most dramatic improvement is in spark cosine similarity, which measures how well the model identifies the core insight. The baseline achieves only 0.221 (near-random alignment) while Run 2b achieves 0.620, a 2.8$\times$ improvement. This suggests the baseline fails to identify the \emph{specific novel contribution}; fine-tuning on reasoning traces teaches the model to ground its output in the key insight.

\begin{figure}[t]
\centering
\includegraphics[width=0.85\textwidth]{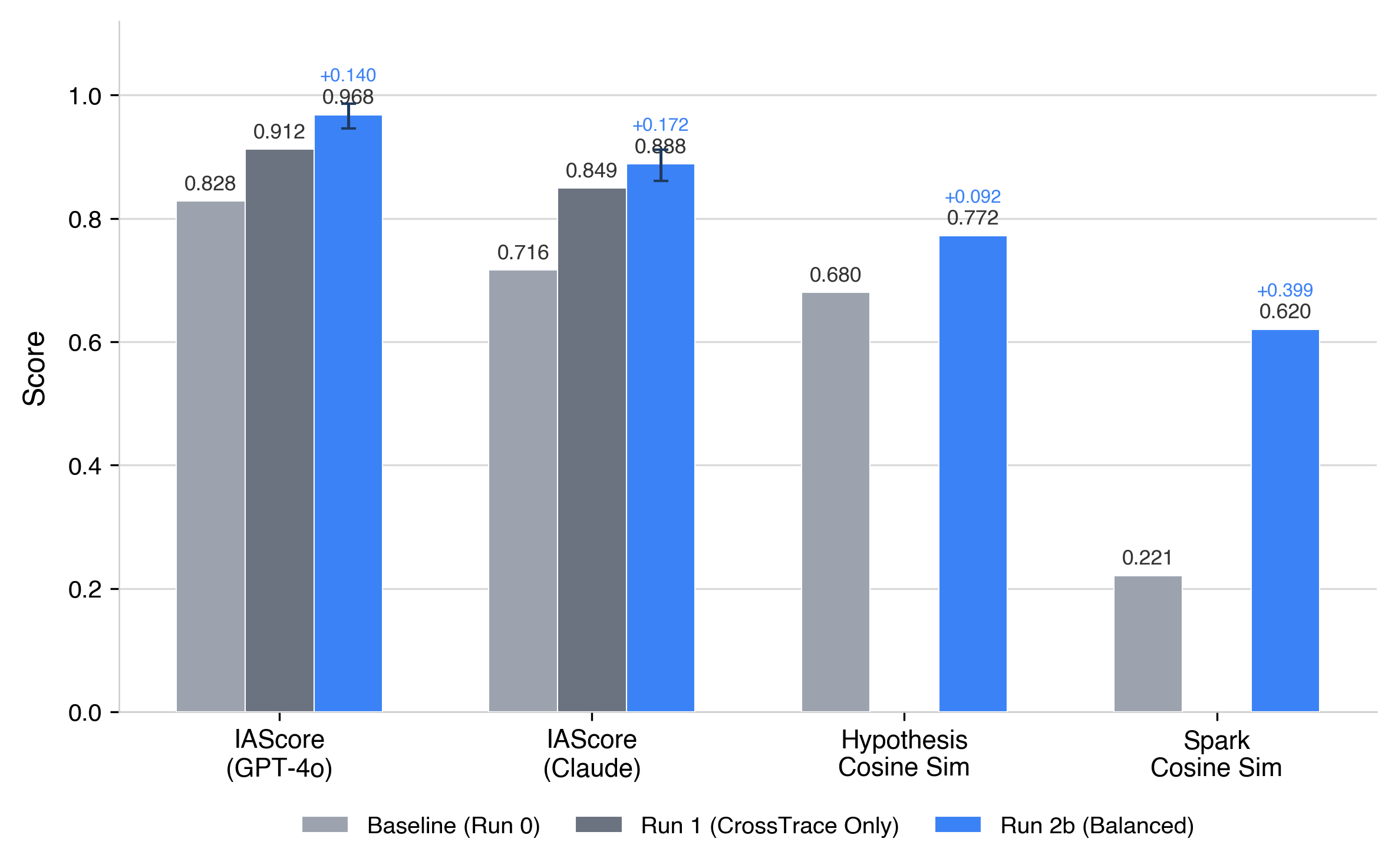}
\caption{Experimental results comparing Baseline, Run 1, and Run 2b across IAScore (both judges) and cosine similarity metrics. Error bars show 95\% confidence intervals.}
\label{fig:results}
\end{figure}

\subsection{Ablation: Domain Transfer vs.\ Scale}
\label{sec:ablation}

To isolate the contribution of cross-domain training from data volume, I trained three matched-size models (1,180 records each, 3 epochs, identical hyperparameters) on different data compositions and evaluated on both test sets. Table~\ref{tab:ablation} presents results.

\begin{table}[t]
\caption{Ablation results isolating domain transfer from scale. All training sets contain exactly 1,180 records. Hypothesis cosine similarity (Hyp Cos) and spark cosine similarity (Spark Cos) are reported against ground-truth references.}
\label{tab:ablation}
\centering
\small
\begin{tabular}{@{}lcccc@{}}
\toprule
Training Data & \multicolumn{2}{c}{CT Test} & \multicolumn{2}{c}{HG Test} \\
\cmidrule(lr){2-3} \cmidrule(lr){4-5}
& Hyp Cos & Spark Cos & Hyp Cos & Spark Cos \\
\midrule
HypoGen-only (1,180 CS) & 0.567 & 0.488 & \textbf{0.769} & \textbf{0.609} \\
CrossTrace-only (1,180 multi-domain) & \textbf{0.673} & \textbf{0.544} & 0.713 & 0.558 \\
Mixed (590 CT + 590 HG) & 0.668 & 0.508 & 0.767 & 0.549 \\
\bottomrule
\end{tabular}
\end{table}

The mixed model, trained on half the in-domain data of each specialist, achieves 99.3\% of CrossTrace-only performance on the CrossTrace test set (0.668 vs.\ 0.673) and 99.7\% of HypoGen-only performance on the HypoGen test set (0.767 vs.\ 0.769). If cross-domain data provided no transferable signal, halving the in-domain training examples should produce measurable degradation; instead, near-parity is maintained. This is consistent with cross-domain transfer: 590 out-of-domain traces compensate for 590 missing in-domain traces, suggesting the model learns domain-general reasoning structure rather than domain-specific patterns.

A secondary finding is that the CrossTrace-only model achieves 0.713 hypothesis cosine similarity on the HypoGen CS-only test set despite never seeing CS-only training data, further supporting transferable reasoning patterns.

\subsection{Human Evaluation of Model Outputs}
\label{sec:humaneval}

To complement automated metrics, I conducted a human evaluation of model-generated hypotheses. Using the mixed ablation model, I generated outputs for 30 stratified test records (13 AI/ML, 11 biomedical, 6 cross-domain) and rated each on four dimensions using a 1--5 scale. Table~\ref{tab:humaneval} presents results.

\begin{table}[t]
\caption{Human evaluation results ($n=30$). Ratings by the first author on model-generated hypotheses from the mixed ablation model, stratified by domain. Scale: 1 (poor) to 5 (excellent).}
\label{tab:humaneval}
\centering
\small
\begin{tabular}{@{}lccccc@{}}
\toprule
Domain & $n$ & Novelty & Usefulness & Soundness & Overall \\
\midrule
Biomedical & 11 & 3.55 & 3.64 & 3.55 & 3.45 \\
AI/ML & 13 & 3.38 & 3.23 & 2.92 & 3.08 \\
Cross-domain & 6 & 2.33 & 2.83 & 3.00 & 2.50 \\
\midrule
\textbf{All} & \textbf{30} & \textbf{3.23} & \textbf{3.30} & \textbf{3.17} & \textbf{3.10} \\
\bottomrule
\end{tabular}
\end{table}

The overall mean of 3.20 across all dimensions indicates that model outputs are consistently above the midpoint, producing hypotheses that are moderately novel, useful, and scientifically sound. Biomedical outputs scored highest across all four dimensions (overall 3.45), consistent with CrossTrace having the most structured grounding traces in that domain. Cross-domain outputs scored lowest (overall 2.50), as expected given the difficulty of reasoning across disciplinary boundaries.

Qualitative analysis of the ratings reveals two failure modes: (1) \emph{domain drift}, where the model shifts the hypothesis away from the paper's specific contribution toward a more generic version; and (2) \emph{mechanism substitution}, where the model replaces the ground truth's specific technical mechanism with a plausible but unfaithful alternative. Both failure modes are more pronounced in AI/ML and cross-domain outputs.

To address the limitation of single-annotator evaluation, I conducted a blinded expert evaluation with three independent reviewers: a clinical medicine specialist, an AI/ML researcher, and a biomedical/computational biology researcher. Each reviewer rated 15 stratified examples (5 per domain) on the same four dimensions without knowledge of the model architecture or training procedure. Table~\ref{tab:experteval} presents aggregate results.

\begin{table}[t]
\caption{Blinded expert evaluation ($n=15$ items $\times$ 3 raters). Reviewers: clinical medicine, AI/ML, biomedical/computational biology. ICC = intraclass correlation coefficient, two-way random, single measures.}
\label{tab:experteval}
\centering
\small
\begin{tabular}{@{}lccccc@{}}
\toprule
Domain & $n$ & Novelty & Usefulness & Soundness & Overall \\
\midrule
Biomedical & 5 & 3.13 & 4.73 & 4.40 & 4.33 \\
AI/ML & 5 & 3.60 & 3.53 & 3.27 & 3.53 \\
Cross-domain & 5 & 3.13 & 4.27 & 3.60 & 3.67 \\
\midrule
\textbf{All} & \textbf{15} & \textbf{3.29} & \textbf{4.18} & \textbf{3.76} & \textbf{3.84} \\
\midrule
ICC(2,1) & --- & 0.495 & 0.557 & 0.929 & 0.447 \\
\bottomrule
\end{tabular}
\end{table}

Expert ratings are consistently higher than the author's initial assessment (Table~\ref{tab:humaneval}), with usefulness (4.18/5) and overall quality (3.84/5) showing the largest gains. Scientific soundness achieves excellent inter-rater agreement (ICC = 0.929), indicating that the three reviewers converge on which reasoning chains are logically valid. Moderate agreement on novelty (0.495) and overall quality (0.447) reflects the inherently subjective nature of these dimensions. Biomedical hypotheses again scored highest (overall 4.33), consistent with the author evaluation.

\section{Analysis}

\subsection{Domain Mixing Analysis}

Comparing Run 1 (1,180 records) and Run 2b (4,720 records) isolates the effect of domain mixing. Run 2b has four times more training data, which could explain its improvement through scale alone. However, two observations suggest domain mixing itself provides signal beyond data volume.

First, Run 2b uses my data at 2$\times$ oversampling. The incremental data is from HypoGen, a different domain with a different trace format. If improvement were purely from volume, narrative-format HypoGen data should provide less benefit per record. Second, Run 2b improves on both the CS-domain HypoGen test set and the multi-domain CrossTrace test set, consistent with cross-domain transfer rather than domain-specific memorization.

\textbf{Confound resolution.} The comparison between Run 1 and Run 2b conflates cross-domain transfer with increased data volume. Section~\ref{sec:ablation} presents a controlled ablation that isolates these factors by training three matched-size models (1,180 records each). The ablation confirms that cross-domain transfer operates independently of scale: a model trained on 590 CrossTrace + 590 HypoGen records achieves near-parity with domain-specialist models trained on 1,180 in-domain records (99.3\% on the CrossTrace test set, 99.7\% on the HypoGen test set).

\subsection{What Transfers Across Domains}

The near-parity of the mixed ablation model with domain specialists raises a natural question: what is being learned that transfers? I propose that grounded reasoning traces teach three domain-independent competencies. First, \emph{problem decomposition}: the ability to break a research gap into an ordered sequence of sub-problems, each building on the previous step. This structure is identical whether the gap concerns protein folding or reinforcement learning. Second, \emph{grounding discipline}: the practice of anchoring each reasoning step to specific evidence rather than generating plausible but unsupported claims. This constraint forces the model to distinguish between what is established and what is hypothesized, a skill that does not depend on domain knowledge. Third, \emph{constraint tracking}: maintaining logical consistency across steps so that the final hypothesis follows from the accumulated evidence rather than appearing de novo. These competencies correspond to the structural features of the Input/Trace/Output schema itself. The schema does not encode domain-specific facts; it encodes the \emph{form} of scientific reasoning. This explains why a model trained on half biomedical and half CS traces can match specialists: the reasoning form is shared, even when the domain content differs.

\subsection{Qualitative Comparison: Baseline vs.\ Fine-Tuned}

To illustrate the practical effect of trace-based training, I compare baseline and fine-tuned outputs on the same biomedical prompt about CKD prediction equity in underserved populations.

\textbf{Baseline (untuned Qwen2.5-7B).} The baseline produces a single unstructured paragraph that restates the prompt context and offers a generic suggestion (``more diverse datasets could improve prediction'') without identifying a specific gap, decomposing the reasoning into steps, or grounding claims in prior work. It does not produce the Input/Trace/Output schema (0\% structural compliance across all 107 test records), making it unusable as a structured hypothesis.

\textbf{Fine-tuned (Run 2b).} The fine-tuned model generates a complete structured trace: it identifies the specific gap (equity-blind validation in existing CKD models), decomposes the reasoning into four grounded steps connecting prior evidence to a novel direction, and produces the spark ``Equitable CKD risk models for diverse cohorts'' (ground truth: ``CKD prediction equity in underserved populations''; spark cosine similarity 0.89). The generated hypothesis specifies health equity as the gap, diverse training data as the approach, and calibration across demographic subgroups as the evaluation criterion.

The key difference is not surface fluency but \emph{reasoning structure}: the fine-tuned model decomposes a research gap into ordered, verifiable steps, while the baseline conflates context and conclusion into a single undifferentiated paragraph. This pattern is consistent across domains. In an AI/ML example on constraint programming, the baseline produces a plausible but vague suggestion about ``improving propagation efficiency,'' while the fine-tuned model identifies the specific limitation (single-constraint reasoning), proposes a concrete mechanism (composite-resource formulation), and predicts a testable consequence (stronger inference through simultaneous propagation).

\section{Limitations}

This work has several limitations that should be considered when interpreting the results.

The LLM-as-judge IAScore used here is not comparable to HypoGen's IdeaMatcher-based metric; absolute scores across frameworks cannot be compared, and I make no claim of surpassing HypoGen's reported results. The CrossTrace test set (107 records) and HypoGen test set (50 records) are relatively small, limiting statistical power for fine-grained per-domain analysis.

All reasoning traces were extracted by a single model (Claude Sonnet 4), introducing systematic biases in which reasoning patterns are identified and how they are articulated. Different extraction models would likely produce different traces from the same papers. Relatedly, the extraction pipeline relies on a single annotator's validation. While human review of 150 stratified records identified no fabricated steps and 99.7\% grounding accuracy, these results may be subject to confirmation bias, and inter-annotator agreement cannot be established from a single reviewer.

My automated evaluation metrics measure alignment between generated and ground-truth hypotheses. The blinded expert evaluation (Table~\ref{tab:experteval}) provides multi-rater validation with moderate to excellent inter-rater agreement, but the sample remains small ($n=15$) and the expert panel limited to three reviewers. A larger-scale evaluation with domain-specific expert panels and downstream task validation (e.g., whether generated hypotheses lead to productive experiments) remains future work.

CrossTrace draws exclusively from English-language preprints on medRxiv, bioRxiv, and arXiv, excluding research published in other languages or in fields not well represented on these servers. Finally, the reasoning traces represent justification narratives (post-hoc rational accounts constructed for publication) rather than actual cognitive discovery processes. This limitation is shared by all work in this space \citep{li2025hypogen, pu2025ideabench, liu2025researchbench}.

\section{Conclusion}

The central finding of this work is that structured, grounded reasoning traces encode transferable reasoning primitives. A controlled ablation on matched-size training sets (Section~\ref{sec:ablation}) shows that a model trained on 590 biomedical and 590 CS traces achieves 99.3\%/99.7\% of domain-specialist performance, confirming that the benefit of trace-based supervision is domain-general rather than an artifact of data volume.

CrossTrace, the dataset underlying these experiments, provides 1,389 grounded scientific reasoning traces spanning biomedical research, AI/ML, and cross-domain work. Fine-tuning Qwen2.5-7B-Instruct on this data yields substantial improvements over the untuned baseline (IAScore 0.968/0.888 with GPT-4o/Claude judges; structural compliance from 0\% to 100\%). A blinded expert evaluation by three independent domain specialists (Section~\ref{sec:humaneval}) validates that fine-tuned models produce hypotheses rated well above the scale midpoint on usefulness (4.18/5), scientific soundness (3.76/5), and overall quality (3.84/5), with excellent inter-rater agreement on soundness (ICC = 0.929).

I plan to extend CrossTrace in three directions: (1) \emph{lineage-aware training} using the schema's \texttt{parent\_record\_id} field to teach models to generate follow-up hypotheses that build on existing work; (2) \emph{scaled expert evaluation} expanding the blinded expert panel to larger samples with domain-specific reviewer pools and downstream task validation; and (3) addressing the failure modes identified in human evaluation, particularly domain drift and mechanism substitution in cross-domain outputs, through targeted data augmentation or constrained decoding strategies.

\section*{Ethics Statement}

This study does not involve human subjects, animal research, or clinical data. All source materials are published, publicly available preprints from arXiv, bioRxiv, and medRxiv. No institutional review board approval was required.

\section*{Acknowledgments}

This work was conducted without external funding. Compute resources were provided through Google Colab Pro (educational license).

\section*{Data Availability}

The CrossTrace dataset (JSONL format), extraction prompts, training configurations, and evaluation scripts are publicly available at \url{https://github.com/andrewbouras/crosstrace}.

\bibliographystyle{plainnat}

\appendix

\section{Extraction Prompt}

The full extraction prompt used for reasoning trace extraction is included in the public repository (\url{https://github.com/andrewbouras/crosstrace}). The prompt contains a detailed output format specification (JSON schema), instructions for reasoning trace granularity (3--6 steps), and grounding requirements. Three few-shot examples are included: a simple gap-fill paper (3 steps), a cross-domain analogy transfer paper (5 steps), and an implicit novelty paper (3 steps, low explicitness).

\section{Training Data Format}

Records are converted to a chat-template format for fine-tuning:

\begin{verbatim}
{"messages": [
  {"role": "system", "content": "You are a scientific reasoning
   assistant. Given a description of the current state of knowledge
   and the conventional assumption being challenged, generate a
   novel hypothesis with step-by-step reasoning."},
  {"role": "user", "content": "Field context: [field_context]\n
   Conventional assumption: [conventional_assumption]\n
   Key prior work: [ref1; ref2; ...]"},
  {"role": "assistant", "content": "Core insight: [spark]\n
   Reasoning:\n1. [step 1]\n2. [step 2]\n...\n
   Hypothesis: [hypothesis]\nApproach: [approach]"}
]}
\end{verbatim}

\section{Database Schema}

The full SQLite schema for the \texttt{novelty\_records} table is available upon request from the corresponding author.

\section{Confidence Interval Computation}

Confidence intervals for IAScore are computed via bootstrap resampling (1,000 iterations) on the test set predictions.

\section{HypoGen Data Integration}

HypoGen records \citep{li2025hypogen} are loaded from the \texttt{UniverseTBD/hypogen-dr1} dataset on HuggingFace. For the balanced training condition (Run 2b), HypoGen's 5,478 training records are randomly subsampled to 2,360, and CrossTrace's 1,180 training records are oversampled 2$\times$ to 2,360, yielding 4,720 total training records.

\end{document}